# Automated Screw Planning for Reduced Pelvic Fractures Based on Statistical Shape Models and Deep Learning

Yang Gao[1], Sutuke Yibulayimu[2], Yanzhen Liu[2], Zian Zhao[3], Yudi Sang[1*]

[1] Beijing Rossum Robot Technology Co., Ltd., Beijing, China.

[2] The Key Laboratory of Biomechanics and Mechanobiology (Beihang University), Ministry of Education, Beijing Advanced Innovation Center for Biomedical Engineering, School of Biological Science and Medical Engineering, Beihang University, Beijing, China.

[3] Beijing 101 High School International Department, Beijing, China.

* Corresponding author: Yudi Sang, email: sangyudi@rossumrobot.cn

## Abstract

Percutaneous iliosacral screw fixation is an important minimally invasive treatment for unstable pelvic fractures. Because the sacroiliac region has complex anatomy and narrow screw corridors, the accuracy and safety of screw placement directly affect surgical outcomes. Accurate and reliable preoperative screw planning is therefore essential to improve surgical success and reduce intraoperative risks. Conventional preoperative planning typically requires surgeons to determine screw trajectories through manual measurements, a labor-intensive process that depends on subjective clinical experience. To address these challenges, we propose a fully automated pipeline for preoperative iliosacral screw planning in patients with pelvic fractures. Using patient-specific three-dimensional anatomy, the pipeline automatically identifies safe screw corridors and generates individualized insertion trajectories to support clinical preoperative planning. Compared with existing methods, the main innovations are as follows: (1) statistical shape models restore normal bone morphology in fractured regions, reducing the influence of fracture-induced deformities on safe corridor identification; (2) screw trajectory landmarks are defined and automatically detected using a neural network, providing a reliable spatial basis for automated screw planning; and (3) a fully automated preoperative iliosacral screw planning pipeline supports multiple anatomical regions and rapidly searches candidate trajectories using these landmarks, enabling simultaneous individualized planning for four key screw placement regions: S1, S2, the superior pubic ramus, and LC-II. We evaluated the proposed pipeline on 200 clinical cases of pelvic fractures. Compared with conventional manual measurements, the safety margin of the safe insertion corridors increased by 2% across the four screw types, the mean planning time decreased by more than 90%, and the clinical acceptance rate reached 95%.

## 1. Introduction

Pelvic fractures are severe orthopedic injuries. Although they account for only 3% of fractures, they are associated with high mortality and disability rates [1]. They typically result from high-energy trauma, such as severe traffic accidents, falls from height, or other high-impact external forces [2]. With global population aging, the incidence of pelvic fractures among older adults is increasing [3]. Osteoporosis and compromised bone structure are common in this population, and even low-energy events, such as falls or slips, can cause pelvic fractures [4].

Percutaneous iliosacral screw fixation is commonly used to treat pelvic fractures. Screw trajectories are planned preoperatively, and intraoperative navigation guides screw insertion to restore pelvic stability. This technique is minimally invasive, involves limited blood loss, and provides good stability. Biomechanical studies have shown that it can restore 85% of pelvic ring stiffness [5]. However, the complex three-dimensional anatomy of the pelvis and its proximity to critical neurovascular structures make percutaneous iliosacral screw insertion technically demanding. In patients with pelvic fractures, displacement and bone defects further alter the original anatomical relationships, complicating the identification of safe screw corridors and increasing the risk of complications associated with screw malposition [6]. Deviations in screw trajectories may injure adjacent neurovascular structures [7]. Moreover, the limited space within safe osseous corridors can lead to cortical perforation or inadequate screw stability, potentially requiring revision surgery in severe cases [8]. Current preoperative screw trajectory planning relies on manual measurements by surgeons and is strongly influenced by their clinical experience and individual practices. Intraoperative changes to the insertion plan can also prolong surgery and increase radiation exposure [9]. An accurate and rapid preoperative trajectory planning method would therefore be valuable in helping clinicians develop patient-specific screw placement plans.

To address these challenges, we propose an end-to-end algorithm for automated pelvic screw planning. Using a three-dimensional pelvic model after fracture reduction, the algorithm automatically plans four screw types—S1, S2, superior pubic ramus, and LC-II—without manual intervention. The main contributions of this study are as follows:

1. We propose a statistical shape model (SSM)-based method for restoring the morphology of fractured pelvic regions and apply it to preoperative pelvic screw planning. To address fracture-related bone defects and altered anatomical relationships, separate SSMs are constructed for the left ilium, right ilium, and sacrum. A shape-fitting objective is formulated, and gradient descent is used to optimize the shape parameters, reconstructing healthy bone morphology from the patient's fractured pelvis. This reduces interference from fracture surfaces during the search for safe screw corridors.

2. Because anatomical localization requirements vary among screw types, we define the landmarks needed to plan iliosacral, superior pubic ramus, and LC-II screws, including entry points, endpoints, and trajectory constraint points. We also develop a lightweight three-dimensional point cloud landmark detection model with PointNet++ as its backbone. The model automatically localizes initial pelvic screw landmarks, providing reliable initial positional constraints for subsequent trajectory optimization through local spatial search.

3. To accommodate differences in osseous anatomy and safe corridor constraints, we design individualized trajectory optimization strategies for S1, S2, superior pubic ramus, and LC-II screws. By combining anatomical constraints, geometric search, and safety distance evaluation, we establish a fully automated preoperative planning pipeline covering multiple anatomical regions and screw types, providing patient-specific screw placement plans for pelvic fracture patients.

## 2. Related Work

As computer technology has become increasingly integrated into medicine, various computer-assisted methods for pelvic screw planning have been proposed. Here, we focus on recent semiautomated and fully automated approaches. Liu et al. [10] extracted fracture surfaces from pelvic fracture reduction results and analyzed the geometric features of their point clouds to determine the number of screws and plan screw implantation, automatically generating screw positions and orientations. However, this method depends on accurate fracture reduction, and its planning accuracy may be affected by complex fractures or missing fragments. Yang et al. [11]introduced the first corridor planning algorithm for an arc-shaped internal fixation screw and used finite element analysis to compare its biomechanical performance with that of conventional fixation screws. Zhang et al. [11] developed a trigonometric algorithm that obtains anatomical parameters from preoperative pelvic CT measurements and uses them to calculate the ideal insertion angle and spatial trajectory of iliosacral screws. Registration has also been widely applied to pelvic screw planning. These approaches typically involve manually delineating approximate entry and exit regions on a template, then registering the template to the patient's pelvis to transfer the screw search regions. Goerres et al. [13] used deformable registration to align an expert-annotated atlas containing the pelvic surface and standard percutaneous screw trajectories with patient CT images, enabling individualized trajectory planning. Intraoperative 3D–2D registration then aligned the CT images with fluoroscopic images to guide screw insertion. Florian Gras et al. [14] constructed a mean pelvic shape model from 523 pelvic specimens. They manually labeled entry and exit regions, mapped them to the target pelvis, and iteratively computed optimal screw trajectories. Both registration-based approaches still require manual adjustment of registration parameters for individual cases, limiting their general applicability. Ju et al. [15] proposed a pelvic screw planning method based on deep learning-based deformable registration. By constructing a pelvic atlas and mapping predefined screw planning information to the patient's pelvis, their method automatically annotated screw search regions. However, the study was conducted only on intact pelves, and its applicability to complex clinical fractures requires further investigation.

## 3. Methods

### 3.1 Overview

Figure 3.1 illustrates the overall workflow of the proposed pelvic screw planning method. All computations are performed on pelvic surface meshes. First, using the constructed statistical shape models of the ilium and sacrum, the patient's fractured sacrum or ilium is used as the input to shape fitting to restore healthy morphology at the fracture site. Next, a pelvic screw landmark detection network identifies the landmarks associated with the four screw types, providing a basis for subsequent trajectory optimization. The network is sufficiently lightweight to accurately locate these landmarks while adding negligible computational overhead to the overall planning process. Finally, a screw trajectory optimization pipeline applies a tailored strategy to each screw type, rapidly and accurately generating screw placement recommendations.

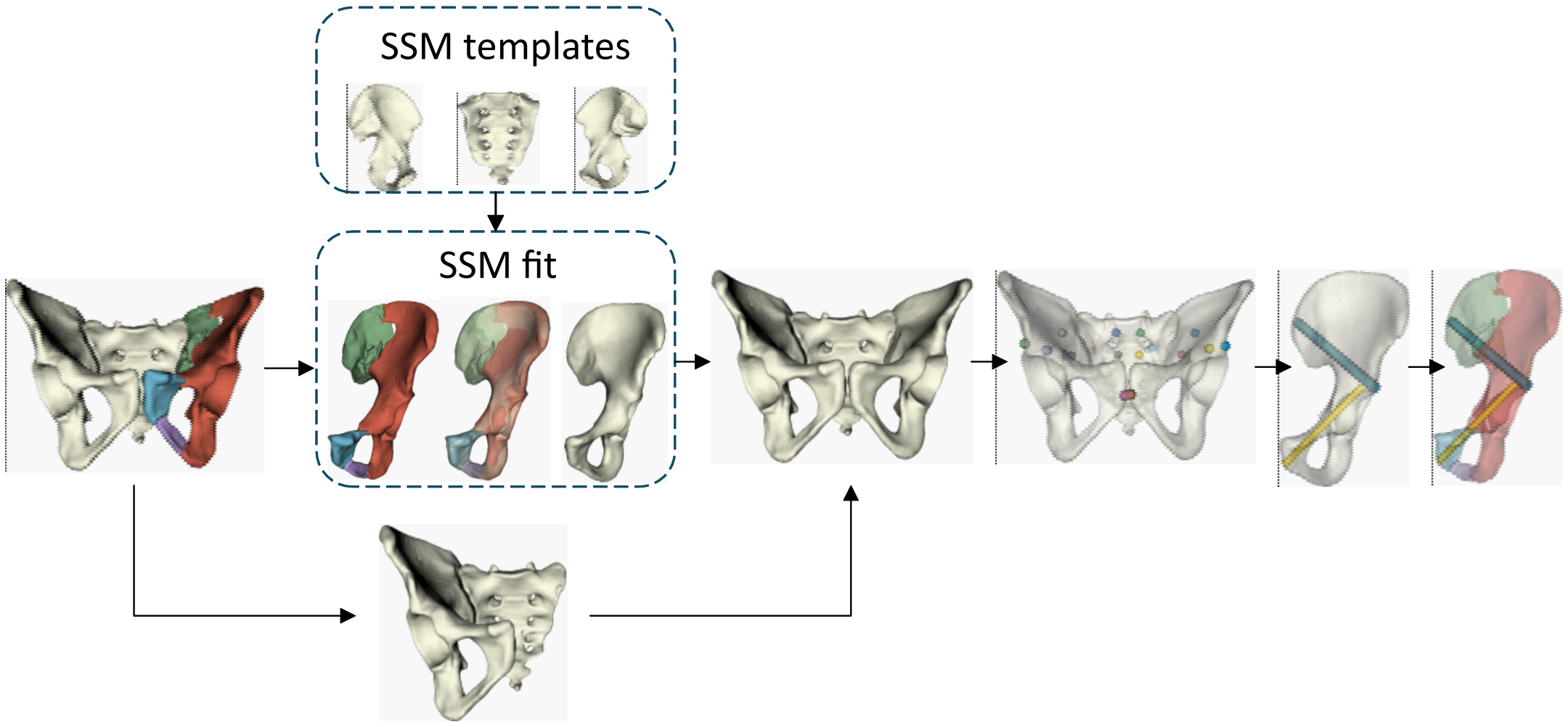

Figure 3.1. Overall workflow.

### 3.2 Construction of Statistical Shape Models

A fracture divides a bone into two or more fragments. Even after reduction, fracture surfaces remain. Because screw trajectory planning is performed on three-dimensional models, these surfaces inevitably affect the trajectory search. We therefore restore the reduced bones to healthy morphology before planning screw trajectories.

A statistical shape model is a statistical approach to shape representation. By modeling a large set of samples from a class of objects, it encodes the dominant modes of variation in a low-dimensional representation that can be used to generate new samples. SSMs are widely used in medical engineering. For brevity, we use the abbreviation SSM throughout the remainder of this paper. In this study, SSMs are used to reconstruct pelvic anatomy after fracture reduction rather than to generate previously unseen samples.

We use the left ilium to illustrate SSM construction and its application to anatomical reconstruction after pelvic fracture reduction. One sample is randomly selected from the left ilium dataset as the initial deformable template, and its resolution is reduced. Surface meshes derived from volumetric data contain large numbers of vertices, which would otherwise impose substantial computational costs on subsequent processing. In this study, each ilium and the sacrum are downsampled to 5,000 vertices to balance the preservation of geometric detail against computational time complexity. An SSM requires the kth point in each sample to represent the same anatomical location. To establish this correspondence, the left ilium template is first coarsely aligned with each remaining sample using the iterative closest point (ICP) algorithm, followed by nonrigid registration using the coherent point drift (CPD) algorithm. This produces a new left ilium dataset in which every sample other than the initial template is obtained by deforming that template. The resulting samples therefore have one-to-one vertex correspondence, and each contains exactly 5,000 vertices.

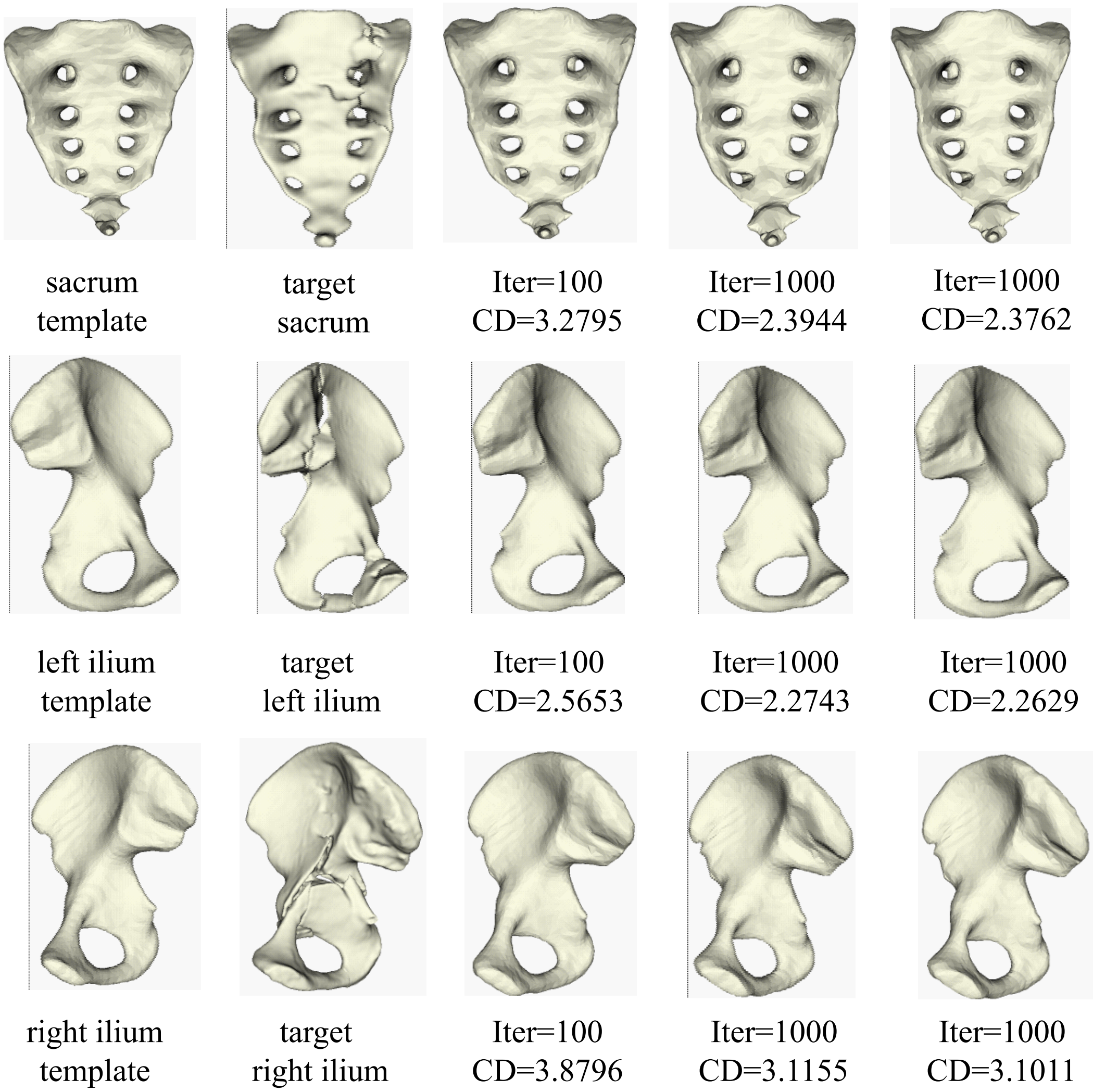

Figure 3.2. Illustration of the SSM fitting process.

Each ilium point cloud is then normalized to achieve spatial alignment and eliminate differences in scale among samples. We compute the mean shape of all left ilium point clouds and apply principal component analysis (PCA) to the dataset, using singular value decomposition (SVD) to obtain the eigenvalues and eigenvectors of the sample covariance matrix. In principle, the mean shape and eigenvectors can represent an arbitrarily shaped left ilium point cloud, parameterized as follows:

$$\text{target shape} = \text{mean shape} + \text{bX} \tag{3-1}$$

Here, b is the vector of weights assigned to the eigenvectors, and X contains the first n eigenvectors; n is set to 40 in this study. In practice, the dataset cannot cover the entire distribution of possible shapes. Moreover, PCA, the core of the SSM, is a dimensionality reduction technique, so the SSM cannot preserve all feature information in the dataset. To improve the model's ability to fit a target sample, we introduce a series of affine transformations, yielding:

$$\text{target shape} = \text{mean shape} + \text{bX} + \text{translation} + \text{rotation} + \text{scale} \tag{3-2}$$

Here, translation, rotation, and scale represent the translation, rotation, and scaling components, respectively. This provides a general expression for a new target shape based on the mean shape of the dataset.

To deform the mean shape toward the target fractured region, we draw inspiration from the training procedure used in deep learning. The fitting function is treated as a simple neural network and optimized using gradient descent. Through iterative optimization, the mean shape progressively approaches the target shape, with the process supervised by a Chamfer distance loss.

### 3.3 Pelvic Point Cloud Landmark Detection Network

We propose defining screw trajectories using screw landmarks, further narrowing the candidate search regions for screw planning, and develop a landmark detection network with PointNet++ as its backbone. For each pelvic point cloud, 16 landmarks are defined according to orthopedic surgeons' clinical practices in percutaneous pelvic screw insertion. These landmarks represent all possible trajectories for the four screw types.

The S1 and S2 screw trajectories are each defined by four landmarks. Taking S1 as an example, two landmarks are located at the center of the S1 osseous corridor. This allows rapid assessment of whether a patient with a sacral fracture is eligible for transverse transsacral screw placement—that is, whether the S1 osseous corridor can accommodate a 7.3 mm-diameter screw. If this criterion is not met, the iliac landmarks are used to plan an oblique screw. The same procedure applies to S2. For LC-II screws, no selection between trajectory types is required. Entry and exit points are defined separately for the left and right ilia. These landmarks are located at the anterior superior iliac spine and the posterior inferior iliac spine. A regular teardrop-shaped profile is visible from a particular viewing angle. Superior pubic ramus screws likewise require no selection between trajectory types. Their landmarks comprise an entry point at the pubic symphysis and a waypoint posterior to the acetabulum. All screw landmarks serve as seed points for the subsequent search for optimal screw trajectories.

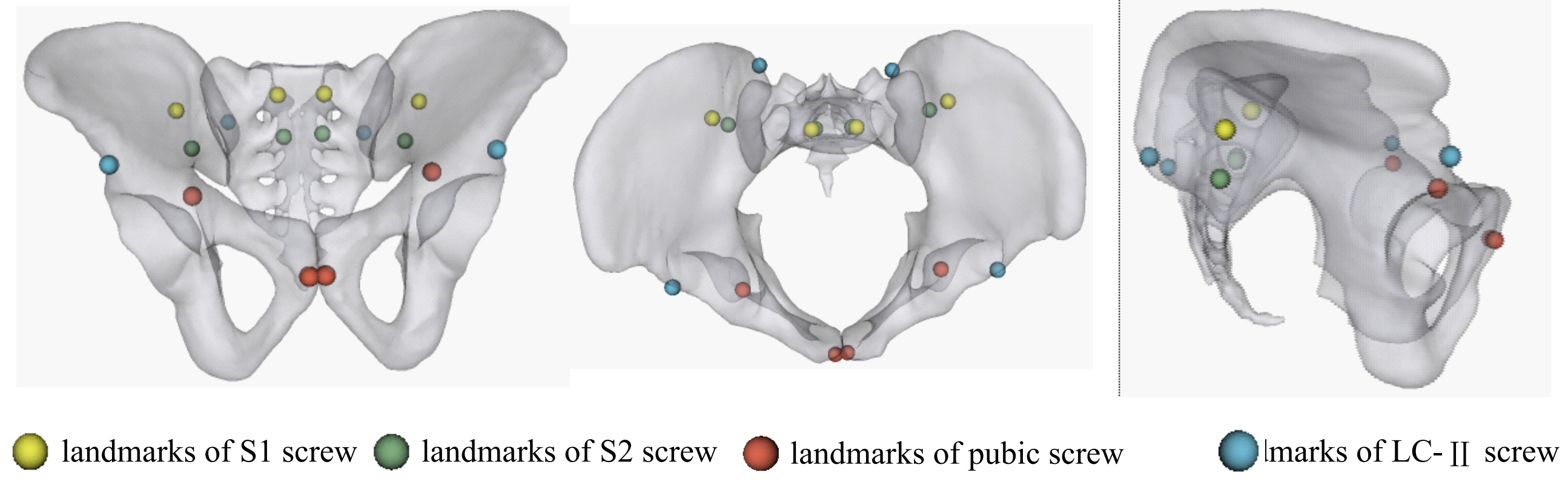

Figure 3.3. Illustration of landmark locations.

PointNet++ is selected as the backbone of the landmark detection network. Compared with PointNet, this backbone learns multiscale features more effectively and offers better generalization for fine-grained tasks and complex scenes. We modify its segmentation head to directly predict heatmaps of pelvic screw landmarks and adopt a coarse-to-fine strategy to further improve localization accuracy.

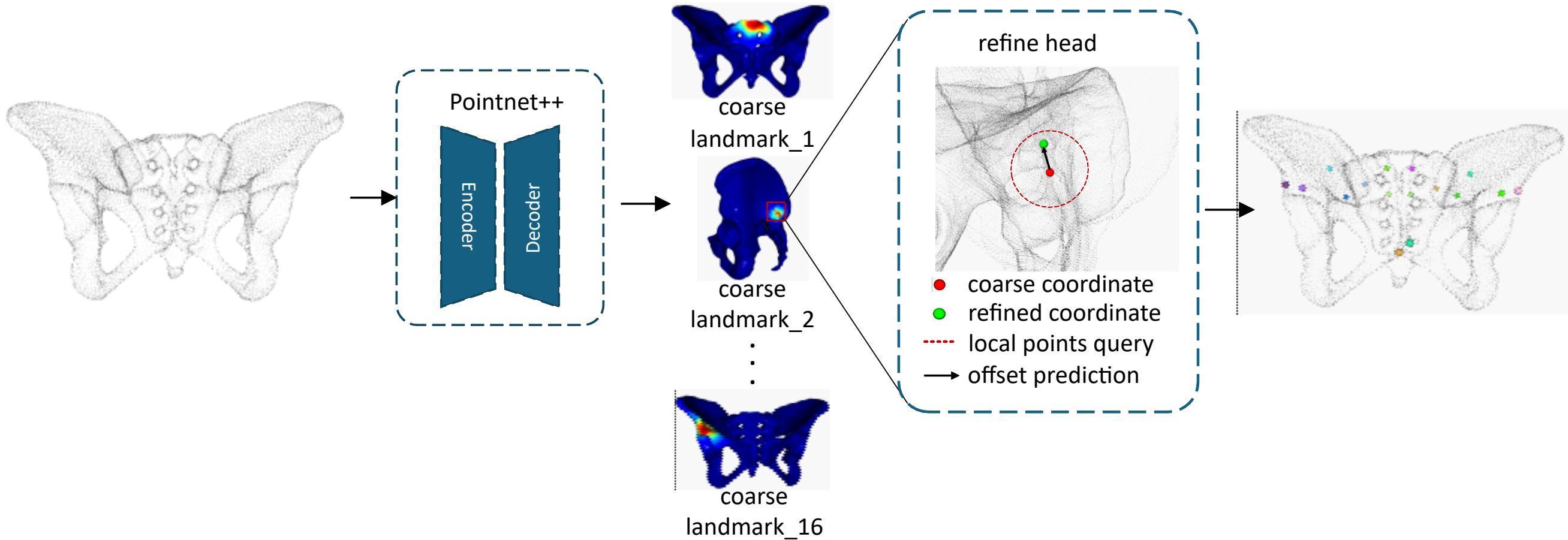

Figure 3.4. Architecture of the landmark detection network.

### 3.4 Screw Trajectory Optimization

Once all screw landmarks have been localized, their initial positions are used to recommend a screw configuration and search for the optimal trajectory.

For S1 and S2 screws, two configurations are commonly used clinically: transverse transsacral placement and oblique unilateral placement. The choice depends on whether the respective S1 or S2 osseous corridor can accommodate a 7.3 mm internal fixation screw, namely, whether the minimum distance from the screw central axis to the inner pelvic surface exceeds 3.65 mm. If sufficient space is available for a 7.3 mm screw, transverse transsacral placement is selected; otherwise, oblique unilateral placement is used. The trajectory optimization algorithm first evaluates the transsacral screw landmarks within the sacrum. Candidate points are uniformly sampled within a specified radius around each landmark in a plane perpendicular to the axis connecting the two landmarks. The sampling radii are 10 mm for S1 and 5 mm for S2, ensuring coverage of their respective osseous corridors. Points outside the pelvic surface are discarded. All pairwise combinations of points from the two candidate sets are then connected to form candidate screw trajectories. The minimum safety distance between each trajectory and the pelvic surface is computed, and trajectories with a minimum safety distance below 3.65 mm are discarded. All candidates satisfying the transsacral placement criterion, with minimum safety distances greater than 3.65 mm, are ranked in descending order of this distance, and the top 20% are retained. PCA is then applied to the set of screw landmarks associated with the retained trajectories. The first principal component defines the insertion direction of the transsacral screw, and the minimum safety distance to the pelvic surface along this direction determines the maximum screw radius that can be accommodated. This approach is intended to avoid errors that may arise from directly adopting the best solution returned by numerical optimization. If no trajectory satisfies the criteria for transsacral placement, an oblique unilateral screw is planned instead.

For superior pubic ramus screws, candidate entry points are generated on the surface of the pubic symphysis. For the other waypoint, candidate points are uniformly sampled within a specified radius around that waypoint in a plane perpendicular to the axis connecting the two landmarks. Pairwise combinations form the set of candidate superior pubic ramus screw trajectories. The minimum safety distance from each trajectory to the inner pelvic surface is computed, and the trajectory with the largest minimum safety distance is selected as optimal.

For LC-II screws, candidate point sets for both landmarks are generated on the iliac surface, and the optimization procedure is the same as that for superior pubic ramus screws.

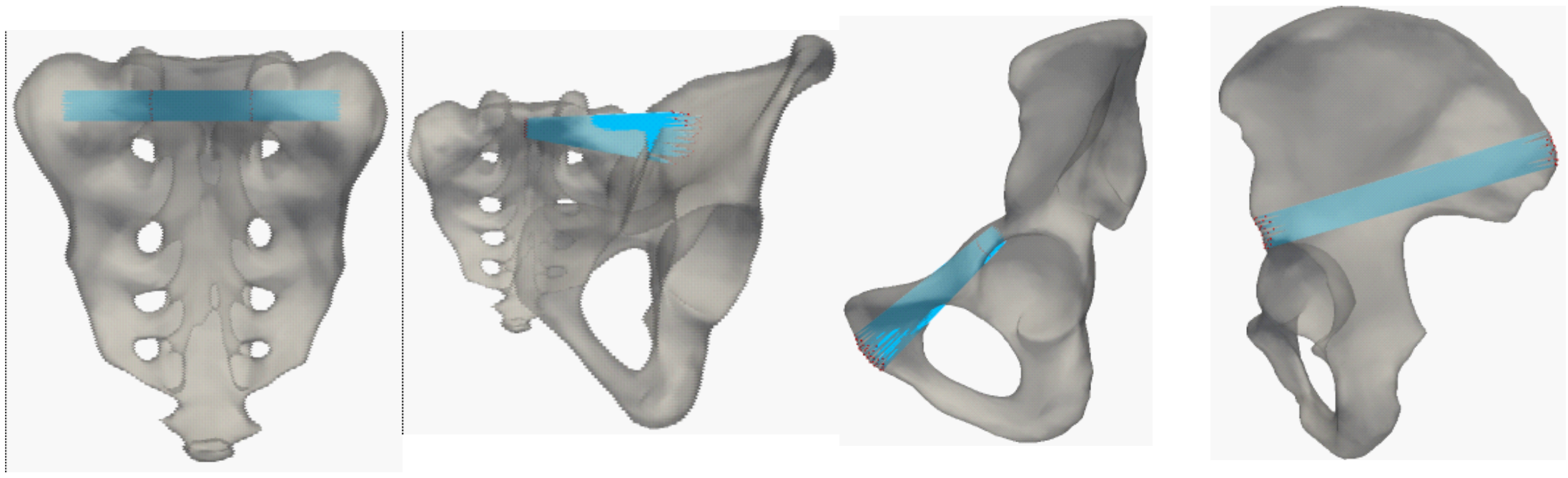

Figure 3.5. Candidate trajectories for different screw types.

## 4. Experiments and Results

### 4.1 Dataset

We assembled a pelvic dataset for algorithm development and testing. All data were obtained from Beijing Jishuitan Hospital and comprised 200 CT scans of intact pelves and 295 CT scans of fractured pelves. All data were de-identified before use, informed consent was obtained from the patients, and experienced orthopedic surgeons manually annotated the scans. For intact pelves, the sacrum and the left and right ilia were labeled separately. For fractured pelves, the annotations were more detailed: in addition to the sacrum and both ilia, each fracture fragment was individually labeled. The intact pelvis data were used to construct the SSMs, while the fractured pelvis data were used to train the screw landmark detection network and test the algorithm. After annotation, the marching cubes algorithm converted the volumetric pelvic labels into surface meshes. The sacrum and the left and right ilia were then screened separately. Samples with substantial missing regions, deformities, previously undetected fractures, or other abnormalities were excluded. The final datasets for SSM construction comprised 175 left ilia, 169 right ilia, and 157 sacra. As the SSM construction procedure was identical for all three regions, the left ilium is used below to illustrate the process.

We also constructed a dataset for model training by manually annotating screw landmarks on 95 fractured pelvis point clouds, using a fixed landmark ordering. Because fracture locations varied, the point distributions differed substantially among samples, making it difficult for the model to converge when trained directly on these data. We therefore generated training data from the annotated dataset using the following strategy. For an individual ilium or sacrum with a fracture, the original point cloud was replaced by a point cloud representing healthy morphology generated through SSM fitting. For an unfractured region, either the original point cloud or the SSM-fitted point cloud could be used. Point clouds for the three regions were combined according to the fracture pattern of each sample to form the training point cloud set. Because the SSM fitting accuracy was high, the landmark labels did not need to be modified for the resulting composite pelvic point clouds. The model was trained for 500 epochs over 6 hours on an NVIDIA A100 GPU. The Adam optimizer was used with a batch size of 8 and an initial learning rate of 0.001, which was gradually reduced during training.

### 4.2 Pelvic Landmark Detection Results

After morphological reconstruction of the fractured regions, accurate localization of screw trajectory landmarks is another essential step. In this section, we evaluate the proposed network on the pelvic screw landmark detection test set. Network predictions are compared with manually annotated landmarks to assess localization accuracy for all 16 landmarks and to compare performance across landmarks associated with different screw types.

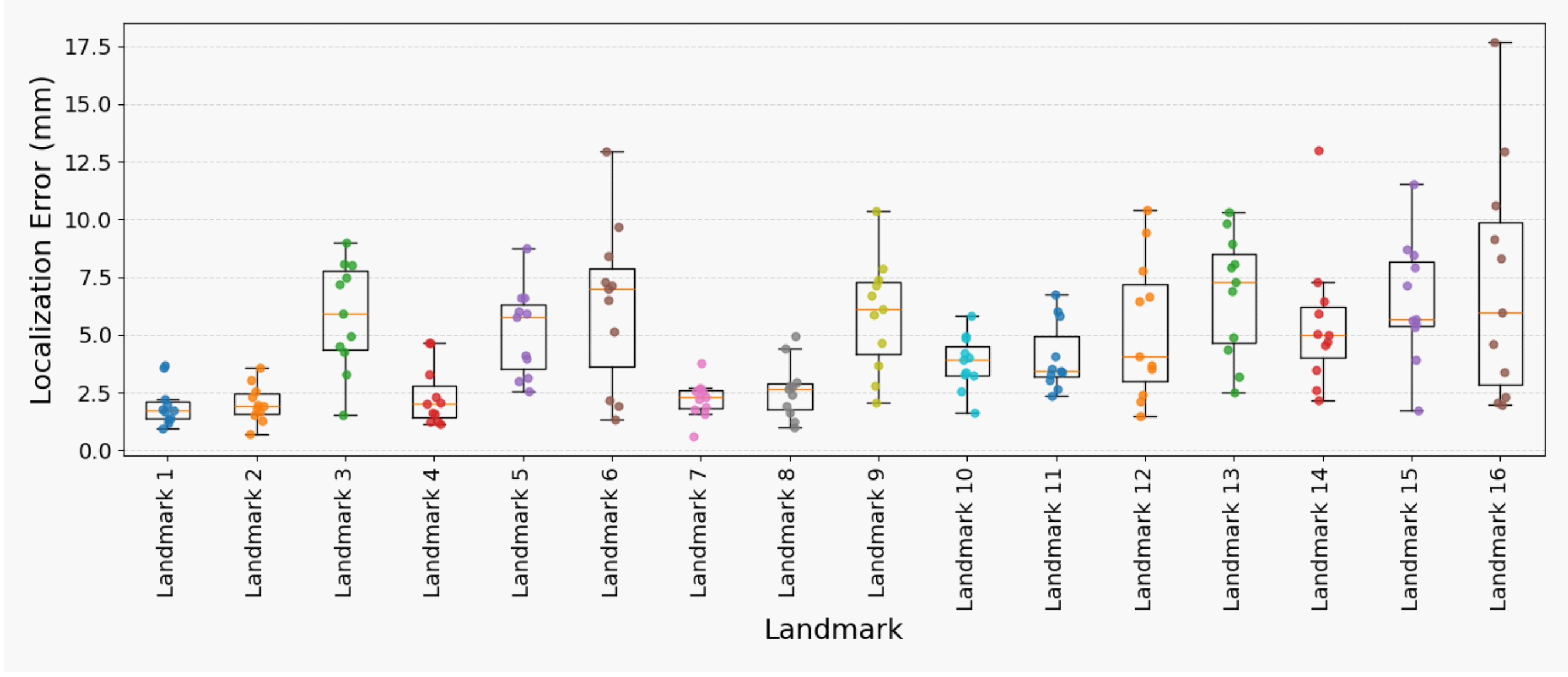

Figure 4.1. Landmark localization errors in the test samples.

Figure 4.1 shows the distributions of localization errors for the 16 landmarks in the test set. Overall, the network consistently localized landmarks across different anatomical regions. All landmarks were successfully predicted, and their predicted positions generally agreed well with the manual annotations. Localization errors varied among landmarks, indicating that the local geometric characteristics of their anatomical regions influenced point cloud landmark detection. Landmarks associated with pronounced changes in local curvature, edges, or distinctive anatomical features were more strongly constrained by the geometry of the surrounding point cloud. The network could therefore learn their local features more readily, resulting in relatively small localization errors. By contrast, some landmarks lay on relatively flat bone surfaces, where similar geometric features extended over a larger area. The lack of distinctive local structural changes limited the network's ability to identify precise landmark positions, resulting in larger errors.

Landmark localization errors do not arise solely from the network itself. Manual annotation also involves a degree of subjectivity, particularly for landmarks without clearly defined anatomical features, whose intended positions may be interpreted differently by different observers. Annotation uncertainty may therefore contribute to the larger prediction errors observed for some landmarks. In addition, reconstructed fracture regions may retain local surface discrepancies relative to the true anatomy, and these reconstruction errors may propagate into subsequent landmark detection.

Overall, although localization accuracy varied among landmarks, the network consistently predicted the positions of all 16 screw trajectory landmarks in the test set, with localization errors remaining within an acceptable range. These results provide a reliable spatial basis for determining screw entry points and endpoints and meet the requirements of subsequent trajectory generation and safety evaluation. Together with the morphological reconstruction results described above, the proposed method establishes a continuous workflow from reconstruction of fractured regions to localization of screw trajectory landmarks, providing a basis for automated screw trajectory planning.

**4.3 Evaluation of Pelvic Screw Planning**

Following morphological reconstruction of the fractured regions and detection of screw trajectory landmarks, screw trajectories were planned using the detected landmarks to assess the overall feasibility of progressing from landmark localization to trajectory generation. Each screw type was planned separately according to its associated landmarks and anatomical region, and the results were visualized in the context of the three-dimensional pelvic anatomy.

Figure 4.2 presents the automatically planned trajectories for different screw types, with cylinders representing the planned screws and their spatial positions. The algorithm generated trajectories from the corresponding landmarks for each screw type. All screws were positioned in the intended anatomical regions and oriented in accordance with the local osseous anatomy. Entry locations and trajectory directions differed clearly among screw types, consistent with their anatomical structures and planning objectives. These results indicate that the detected landmarks provide effective spatial constraints for subsequent trajectory generation.

In three dimensions, the planned screws were generally contained within the corresponding osseous structures, and their axes aligned well with the surrounding bone anatomy. For screws traversing long osseous corridors, the planned trajectories followed the principal direction of the local bony structure. For screws with different entry regions, the algorithm adjusted the insertion direction and spatial orientation according to the associated landmark positions. The generated trajectories were therefore anatomically plausible in both position and direction.

Together with the landmark detection results in the preceding section, these findings demonstrate that landmark localization provides an important basis for screw trajectory planning. The predicted landmarks supply the start and end positions and directional constraints required for different screw types. Subsequent spatial trajectory search and screening complete a continuous pipeline from morphological reconstruction of fractured regions through landmark localization to screw trajectory planning. These visualizations show that the proposed method can generate plans with well-defined anatomical positions and spatial orientations for different pelvic screw types, providing a basis for subsequent screw safety evaluation and clinical planning assistance.

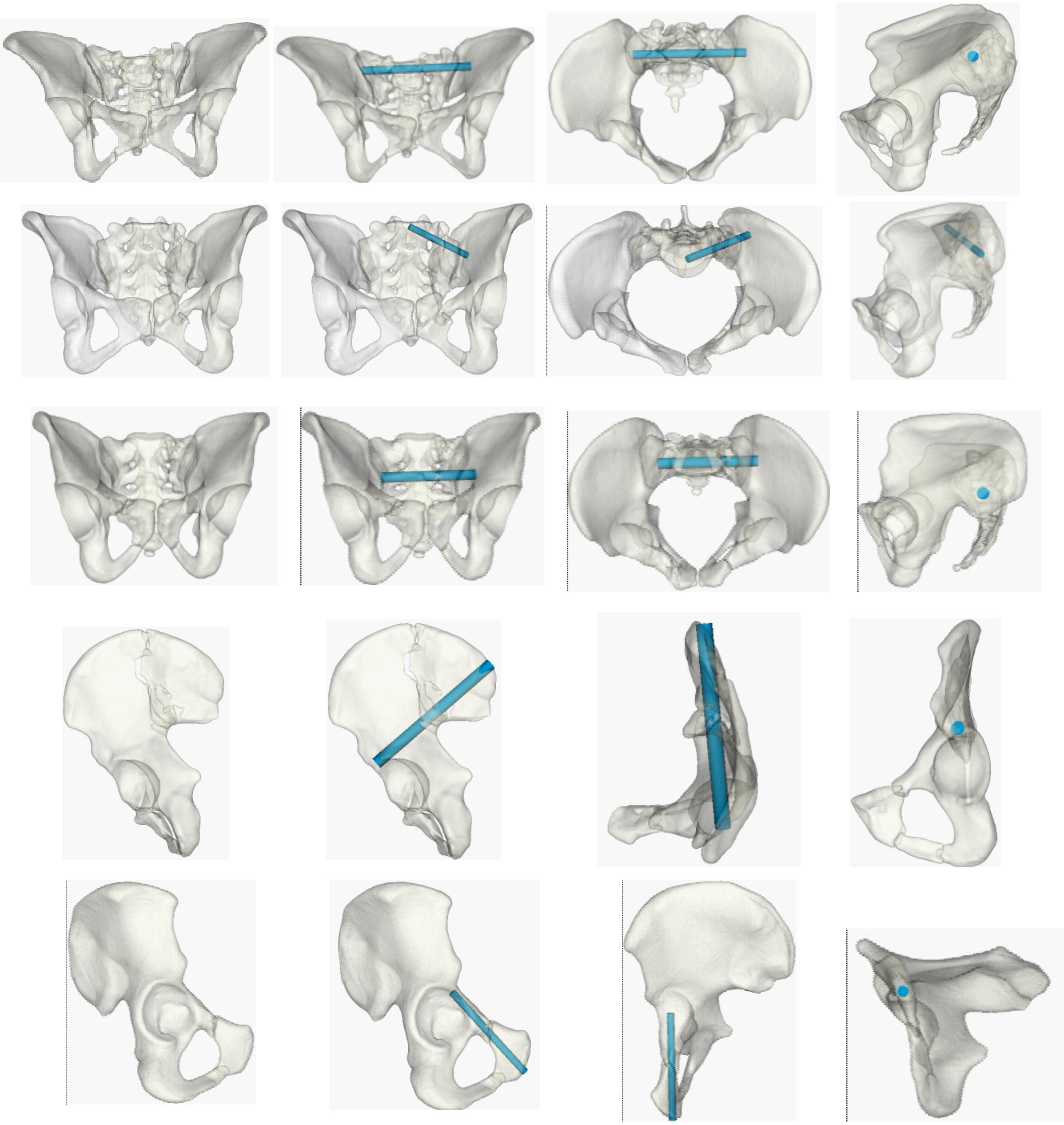

Figure 4.2. Visualization of planning results for different screw types.

Planning safety was evaluated using the minimum distance between the screw and the surrounding cortical bone. A minimum distance of at least 1 mm was classified as excellent, a distance of 0–1 mm as acceptable, and a distance below 0 mm as a failure. The screw planning results showed high success rates across different pelvic screw corridors. The results in Table 4.1 likewise indicate high planning success rates across the evaluated corridors. For S1 and S2 screws, the commonly used 7.3mm diameter was adopted as the evaluation standard, and both achieved overall acceptance rates of at least 98%. LC-II screws were also evaluated using a 7.3 mm diameter and achieved a 100% overall acceptance rate, with no planning failures. Because the superior pubic ramus corridor is generally narrower than the other corridors, particularly in women, children, and older adults with osteoporosis, screw diameters of 4.5 mm and 6.5 mm were used for evaluation. With 4.5 mm screws, the proportion of excellent plans was at least 90%, and the overall acceptance rate reached 100%, with no failures. With 6.5 mm screws, the overall acceptance rate remained at least 99%. These results indicate that the proposed method can effectively generate safe screw insertion trajectories at different anatomical sites.

Table 4.1. Acceptance rates of screw planning results at different anatomical sites.

| Anatomical site | Reference screw diameter / mm | Excellent (≥1 mm) | Acceptable (0–1 mm) | Failure (<0 mm) |
|---|---|---|---|---|
| S1 | 7.3 | 96.98% | 1.51% | 1.51% |
| S2 | 7.3 | 92.46% | 7.54% | 0.00% |
| Left LC-II | 7.3 | 99.50% | 0.50% | 0.00% |
| Right LC-II | 7.3 | 99.50% | 0.50% | 0.00% |
| Left superior pubic ramus | 4.5 | 93.97% | 6.03% | 0.00% |
| | 6.5 | 80.40% | 19.10% | 0.50% |
| Right superior pubic ramus | 4.5 | 97.49% | 2.51% | 0.00% |
| | 6.5 | 78.39% | 21.11% | 0.50% |

## 5. Discussion

We proposed a fully automated pelvic screw planning method based on deep learning and SSMs and evaluated it on 200 cases of pelvic fractures. The experimental results showed that overall acceptance rates were at least 98% at all four anatomical sites, indicating considerable potential for clinical application.

Compared with previous studies, a key strength of this work is that the entire pipeline was developed using clinical data from fractured pelves. Intact pelves have complete, continuous osseous structures and relatively regular anatomy, so screw planning methods developed using intact pelvis data typically produce consistent results. In actual pelvic fracture patients, however, fractures cause displacement, deformation, and even loss of local bone structures, introducing greater uncertainty into trajectory planning. The applicability of methods developed using intact pelves to real clinical fracture data therefore requires careful consideration. This study focuses on pelvic fracture cases and performs automated landmark detection and trajectory planning after fracture reduction and morphological reconstruction. The proposed method was thus evaluated under complex anatomical conditions that more closely reflect clinical practice.

Although the proposed pelvic screw planning method was validated on 200 clinical cases and achieved favorable results, this sample size remains insufficient to cover the full range of clinical presentations, age groups, and fracture mechanisms. Future work should validate the algorithm in larger clinical datasets and more representative patient populations.

## 6. Conclusion

We developed a fully automated pelvic screw planning method that integrates conventional data analysis with deep learning. Taking triangulated meshes of reduced pelvic bone fragments as input, the method identifies optimal screw corridors at four sites: S1, S2, LC-II, and the superior pubic ramus. It provides rapid planning with highly reliable results and requires no manual intervention. Developed and validated entirely using real clinical fracture data, the method has substantial value for supporting clinical decision-making.